\documentclass[conference]{IEEEtran}
\usepackage{hyperref}

\usepackage{amsmath}
\usepackage{tikz}
\usetikzlibrary{matrix,calc}
\usepackage{kbordermatrix}% http://www.hss.caltech.edu/~kcb/TeX/kbordermatrix.sty
\ifCLASSINFOpdf
\else
\fi
\begin{document}
%
% paper title
% Titles are generally capitalized except for words such as a, an, and, as,
% at, but, by, for, in, nor, of, on, or, the, to and up, which are usually
% not capitalized unless they are the first or last word of the title.
% Linebreaks \\ can be used within to get better formatting as desired.
% Do not put math or special symbols in the title.
\title{Scalable Deep Traffic Flow Neural Networks for Urban Traffic Congestion Prediction}

% author names and affiliations
% use a multiple column layout for up to three different
% affiliations

% conference papers do not typically use \thanks and this command
% is locked out in conference mode. If really needed, such as for
% the acknowledgment of grants, issue a \IEEEoverridecommandlockouts
% after \documentclass

% for over three affiliations, or if they all won't fit within the width
% of the page, use this alternative format:
% 
\author{\IEEEauthorblockN{Mohammadhani Fouladgar\IEEEauthorrefmark{1},
Mostafa Parchami\IEEEauthorrefmark{1},
Ramez Elmasri\IEEEauthorrefmark{2} and 
Amir Ghaderi\IEEEauthorrefmark{1}}
\IEEEauthorblockA{\IEEEauthorrefmark{1}Department of Computer Science and Engineering\\
University of Texas at Arlington\\ Email: \textit{firstname.lastname@mavs.uta.edu}}
\IEEEauthorblockA{\IEEEauthorrefmark{2}Department of Computer Science and Engineering\\
University of Texas at Arlington\\ Email: \textit{elmasri@uta.edu}}
}

% use for special paper notices
%\IEEEspecialpapernotice{(Invited Paper)}

% make the title area
\maketitle

% As a general rule, do not put math, special symbols or citations
% in the abstract
\begin{abstract}
Tracking congestion throughout the network road is a critical component of Intelligent transportation network management systems. Understanding how the traffic flows and short-term prediction of congestion occurrence due to rush-hour or incidents can be beneficial to such systems to effectively manage and direct the traffic to the most appropriate detours. Many of the current traffic flow prediction systems are designed by utilizing a central processing component where the prediction is carried out through aggregation of the information gathered from all measuring stations. However, centralized systems are not scalable and fail provide real-time feedback to the system whereas in a decentralized scheme, each node is responsible to predict its own short-term congestion based on the local current measurements in neighboring nodes.

We propose a decentralized deep learning-based method where each node accurately predicts its own congestion state in real-time based on the congestion state of the neighboring stations. Moreover, historical data from the deployment site is not required, which makes the proposed method more suitable for newly installed stations. In order to achieve higher performance, we introduce a regularized euclidean loss function that favors high congestion samples over low congestion samples to avoid the impact of the unbalanced training dataset. A novel dataset for this purpose is designed based on the traffic data obtained from traffic control stations in northern California. Extensive experiments conducted on the designed benchmark reflect a successful congestion prediction.
\end{abstract}

% no keywords

% For peer review papers, you can put extra information on the cover
% page as needed:
% \ifCLASSOPTIONpeerreview
% \begin{center} \bfseries EDICS Category: 3-BBND \end{center}
% \fi
%
% For peerreview papers, this IEEEtran command inserts a page break and
% creates the second title. It will be ignored for other modes.
\IEEEpeerreviewmaketitle

\section{Introduction and related work}
Traffic congestion leads to extra gas emissions and
low transportation efficiency, and it wastes a lot of individuals' time and a hunge amount of fuel. Diagnosing congestion and building a pattern for predicting traffic congestion has been regarded as one the most important issues as it can lead to informal decisions on the routes that motorists take, and on expanding road networks and public transport. Research to predict traffic congested spots, especially in urban areas is thus very important.Typcally, congestion prediction can be used in Advanced Traffic Management Systems (ATMSs) and Advanced Traveller Information Systems in order to develope proactive traffic control strategies and real-time route guidance.\cite{zheng2006short}

In the last decades, concepts of traffic bottleneck and congestion propagation have been considered in many studies. Although most of these originate from Civil Engineering and Urban Transportation studies, the advent of super powerful computers and complex algorithms, traffic management and traffic flow prediction to become an interdisciplinary study.
 
In this regard, there have been various efforts to predict short-term traffic flow prediction, including mathematical equations \cite{zhang2003short,vanajakshi2009improved}, simulation techniques \cite{juri2007integrated}, or statistical and regression approaches. However, traffic flow is based on individuals' decisions, which more likely can be modeled by Artificial Neural Network the best. In other words, traffic flows are made by individuals' decisions based on their knowledge about currenct traffic and their experiences about past traffic flows, which can be modeled by Artificial Neural Network. Using Neural Network for modeling traffic flow and congestion prediction came to the picture in 1993 in \cite{dougherty1993use}. This work propose a network consisting of one \textit{input} layer, one \textit{hidden}, and one \textit{output} layer. Although this structure was proven to perform well in many applications for predicting traffic flow and travel time and estimation, it was not efficient in lots of other, because of the simple structure. Therefore, some research uses a Neural Network, initially, to extract traffic flow patters (clustering), and then based on each pattern, they come up with a proper model to predict traffic flow \cite{rilett2001direct, yin2002urban, kuchipudi2003development}. In this trend, \cite{zheng2006short} different predictors have different performance for various particular time periods. In other words, each predictor can have a super performance only in a particular time period. Therefore, they combined several predictors together as module to have a better performance for longer time periods.

The data regarding Traffic Flow and Traffic Congestion are two instances of Spatio-temporal data. They embady a location (Spatial Feature) and a time (Temporal feature). Besides, as we already mentioned, traffic flow and traffic congestion are based on human actions \cite{baccouche2011sequential}. In \cite{ma2015large}, the authors propose a fully automatic deep model for human-action-based spatio-temporal data. This model first utilizes Convolutional Neural Network model (CNN) to learn the spatio-temporal features.  Then, in the second part of this model, they use the output of the first step to train a recurrent neural network model (RNN) in order to classify the entire sequence. \cite{ma2015large} does not mention traffic issues as one of the possible applications of their work, however it seems promising to make some model, which is inspired by their model, to predict traffic flow and congestion. 

In 2015, \cite{ma2015large} Deep Learning theory was put into practice for large-scale congestion prediction. To this end, they utilized Restricted Boltzmann Machine \cite{hinton2006reducing} and Recurrent Neural Network \cite{goller1996learning} to model and predict the traffic congestion. In order to do this, they convert all the speed data of Taxis in Ningbo, China to binary values (i.e. the speed more than a threshold is 1, otherwise it is 0), and then call these values \textit{Congestion Conditions}. Therefore, the network congestion condition data will be a matrix as follows:
\[
\begin{bmatrix}
   
    C_{1}^{1} & C_{1}^{2} & C_{1}^{3} & \dots  & C_{1}^{T} \\
    C_{2}^{1} & C_{2}^{2} & C_{2}^{3} & \dots  & C_{2}^{T} \\
    \vdots & \vdots & \vdots & \ddots & \vdots \\
    C_{N}^{1} & C_{N}^{2} & C_{N}^{3} & \dots  & C_{N}^{T}

\end{bmatrix}     
\]
Each element in the matrix indicates congestion condition in a specific point at a specific time slot. Therefore, $C_{n}^{t}$ represents the congestion condition on the \textit{n}th point of the traffic network at \textit{t}th time slot (The Network has \textit{N} point). Give this matrix to the model presented in \cite{ma2015large}, the result will be the predicted traffic condition for each point at \textit{T+1}.
\[
\begin{bmatrix}
    
    C_{1}^{1} & C_{1}^{2} & C_{1}^{3} & \dots  & C_{1}^{T} \\
    C_{2}^{1} & C_{2}^{2} & C_{2}^{3} & \dots  & C_{2}^{T} \\
    \vdots & \vdots & \vdots & \ddots & \vdots \\
    C_{N}^{1} & C_{N}^{2} & C_{N}^{3} & \dots  & C_{N}^{T}
\end{bmatrix}
=====\Rightarrow
\begin{bmatrix}
   C_{1}^{T+1}  \\
    C_{2}^{T+1} \\
    \vdots  \\
    C_{N}^{T+1} 
\end{bmatrix}
\] 
%\begin{tikzpicture}%[every node/.style={anchor=north %east,fill=white,minimum width=1.4cm,minimum height=7mm}]
%\matrix (mA) [draw,matrix of math nodes]
%{
% C_{1}^{1} & C_{1}^{2} & C_{1}^{3} & \dots  & C_{1}^{T} \\
%    C_{2}^{1} & C_{2}^{2} & C_{2}^{3} & \dots  & C_{2}^{T} \\
%    \vdots & \vdots & \vdots & \ddots & \vdots \\
%    C_{N}^{1} & C_{N}^{2} & C_{N}^{3} & \dots  & C_{N}^{T}\\
%};
%
%\matrix (mB) [draw,matrix of math nodes] at ($(mA.south east)+(1.5,0.7)$)
%{
%(1,1,3) & (1,1,3) & (1,1,3) & (1,1,3) \\
%(1,1,3) & (1,1,3) & (1,1,3) & (1,1,3) \\
%(1,1,3) & (1,1,3) & (1,1,3) & (1,1,3) \\
%(1,1,3) & (1,1,3) & (1,1,3) & (1,1,3) \\
%};
%
%\matrix (mC) [draw,matrix of math nodes] at ($(mB.south east)+(1.5,0.7)%$)
%{
%(1,1,3) & (1,1,3) & (1,1,3) & (1,1,3) \\
%(1,1,3) & (1,1,3) & (1,1,3) & (1,1,3) \\
%(1,1,3) & (1,1,3) & (1,1,3) & (1,1,3) \\
%(1,1,3) & (1,1,3) & (1,1,3) & (1,1,3) \\
%};
%
%\draw[dashed](mA.north east)--(mC.north east);
%\draw[dashed](mA.north west)--(mC.north west);
%\draw[dashed](mA.south east)--(mC.south east);
%\end{tikzpicture}
Although \cite{ma2015large} presented a good performance for predicting traffic condition, it has some drawbacks:

\begin{itemize}
\item The traffic condition is limited in either \textit{Congested} or \textit{Not-Congested} (1 or 0). However, in real applications, we usually need a \textit{range} of values (or colors in case of Map) to show amount of traffic flow.
\item The traffic condition is set based on a specific threshold (for example 20 km/h). If the average speed is less than the threshold the traffic condition will be set as \textit{congested}, otherwise it will be \textit{Not-congested}. Nevertheless, having a specific threshold for the whole network is inappropriate. Rather, the traffic condition is supposed to be set based on the ratio of average speed of vehicles to possible max speed (Speed limit).
\item In the model presented in \cite{ma2015large}, authors did not consider any order for Network points as the input (the rows of the matrix). However, the spatial influence of adjacent network  points should be taken into consideration.
\end{itemize}

In this paper, we try to predict the traffic flow of Traffic Network points, where we do not have any historical data about them, based on the traffic patterns of Traffic Network points. Therefore, our contributions are as follows:
\begin{enumerate}
\item We formally define the traffic flows prediction concepts.
\item We introduce a normalized data representation, which can be used in Neural Network algorithms, or other methods. 
\item We present a Deep Convolutional Network, which can be able to learn traffic flow of different traffic points.
\item Then, we present a Recurrent Neural Network, which, apart from its structure, can do the same as Convolutional Network.
\item Both of these models are able to predict \textit{n}-level traffic prediction for different points of the traffic (e.g. Quiet, light traffic, heavy traffic, congested, etc.)
\item They also put up the predicted average speeds on different points of the traffic network based on the speed limits in that point (e.g. 0.65 of speed limit).
\item Then, we present a Recurrent Neural Network
\end{enumerate}

In the rest of this paper, we start our work by some preliminaries in section \ref{prelim}. We define all the main traffic flow concepts and then bring up the problem we are going to solve. We also, introduce two deep network model, and describe their structures broadly. In section \ref{method}, we explain our models and methods in more details. Then, we experimentally evaluate our proposed prediction models and compare them with more simple models in section \ref{exprimental}.  
\section{Preliminaries}\label{prelim}
In this section, we give formal definitions of traffic flows prediction problem in subsection \ref{formaldef}. Then in subsection \ref{problem}, we formally introduce the problem presented in this work.  
\subsection{Formal Definitions} \label{formaldef}
 A traffic Network comprises a set of roads, as well as set of junctions. Junctions can be intersections of streets, exit-entrance of highways, roundabouts, beginning-end point of a road, U-turns, etc. In the subject of traffic flows, we can consider junctions as the main points of the network, because these points can be major factors of changing the traffic flows. By way of explanation, typically a traffic flow may not change significantly between two junctions, but it may change because of traffic lights, exit-entrance, and so on. Consequently, almost all of the traffic Network points and traffic sensors are installed in junctions.  

\textbf{Definition 1 (Network Points).} In this study, we represent a road (streets, highways, etc.) by \textit{N} points based on the junctions on that road. Each of these points may indicate the traffic condition (for example congested) on that point. Consequently, the whole Traffic Network Condition can be presented by set of all junctions on that Network, which are called \textit{Network Points} and denoted by $\mathcal {N}$. It is worth mentioning that each Network point has spatial interaction with adjacent Network points, and traffic flow conditions of a point may get/have influence from/on the adjacent points.   

\textbf{Definition 2 (In-flow sequence).} Assume $\mathcal {S}$ is a Network point and $L_{1}$, $L_{2}$, ..., and $L_{n}$ are adjacent points on the traffic network, which have flow to $\mathcal {S}$, such that $L_{1}$ is the closest point to $\mathcal {S}$, and $L_{n}$ is the farthest. The In-flow sequence of $\mathcal {S}$ is denoted as In($\mathcal {S}$).

In($\mathcal {S}$) :
$L_{n}$ $\rightarrow$ $L_{n-1}$ $\rightarrow$ ... $\rightarrow$ $L_{2}$ $\rightarrow$ $L_{1}$ $\rightarrow$ $\mathcal{S}$

\textbf{Definition 3 (Out-flow sequence).} Assume $\mathcal {S}$ is a Network point and $R_{1}$, $R_{2}$, ..., and $R_{m}$ are adjacent points on the traffic network, which $\mathcal {S}$ has flow to them, such that $R_{1}$ is the closest point to $\mathcal {S}$, and $R_{m}$ is the farthest. The Out-flow sequence of $\mathcal {S}$ are denoted as Out($\mathcal {S}$).

Out($\mathcal {S}$) : $\mathcal{S}$ $\rightarrow$ $R_{1}$ $\rightarrow$ $R_{2}$ $\rightarrow$ ... $\rightarrow$ $R_{m-1}$ $\rightarrow$ $R_{m}$ 

\textbf{Definition 4 (Point snapshot).} Assume $\mathcal {S}$ is a Network point and $L_{1}$, $L_{2}$, ..., and $L_{n}$ are In($\mathcal {S}$), and $R_{1}$, $R_{2}$, ..., and $R_{m}$ are Out($\mathcal {S}$), such that:

$L_{n}$ $\rightarrow$ ... $\rightarrow$ $L_{2}$ $\rightarrow$ $L_{1}$ $\rightarrow$ $\mathcal{S}$ $\rightarrow$ $R_{1}$ $\rightarrow$ $R_{2}$ $\rightarrow$ ... $\rightarrow$ $R_{m}$ 

Point snapshot at time \textit{t}, denoted as Snapshot($\mathcal {S}$, \textit{t}), is the \textit{Traffic Condition} of$\linebreak$ [  $L_{n}$, ... , $L_{2}$, $L_{2}$, $\mathcal{S}$, $R_{1}$, $R_{2}$, ..., $R_{m}$ ] at time series of [\textit{t-$\delta$}, ..., \textit{t-1}, \textit{t}]. This time series indicates a sequence of the last $\delta$ time points with a specific time interval between each two consecutive time points (for instance, 20 minutes) . Formally, Snapshot($\mathcal{S}$,t) is defined as follows:
\[
\begin{array}{lc}
 & \kbordermatrix{\text{}&t-\delta&\ldots&t-1&t\cr
                L_n&c_{t-\delta}^{L_n} &  \ldots  & c_{t-1}^{L_n} & c_{t}^{L_n}\cr
\\      
                L_{n-1}&c_{t-\delta}^{L_{n-1}} &  \ldots  & c_{t-1}^{L_{n-1}} & c_{t}^{L_{n-1}}\cr
                \vdots& \vdots & \ddots & \vdots & \vdots\cr
				L_{1}& c_{t-\delta}^{L_{1}} &  \ldots  & c_{t-1}^{L_{1}} & c_{t}^{L_{1}}\cr                
  \\              \mathcal{S}&c_{t-\delta}^{\mathcal{S}} &  \ldots  & c_{t-1}^{\mathcal{S}} & c_{t}^{\mathcal{S}}\cr
	\\			R_{1}& c_{t-\delta}^{R_{1}} &  \ldots  & c_{t-1}^{R_{1}} & c_{t}^{R_{1}}\cr  
                \vdots& \vdots & \vdots & \ddots & \vdots\cr
				R_{m-1}&c_{t-\delta}^{R_{m-1}} &  \ldots  & c_{t-1}^{R_{m-1}} & c_{t}^{R_{m-1}}\cr
          \\      R_m& c_{t-\delta}^{R_{m}} &  \ldots  & c_{t-1}^{R_{m}} & c_{t}^{R_{m}}} \\[25pt]
\end{array}
\]

Where $c_{\tau}^{\sigma}$ indicates \textit{traffic condition} of point $\sigma$ at time $\tau$. \textit{Traffic condition} is a value between 0 and 1 (0 $\leq$ $c_{\tau}^{\sigma}$ $\leq$ 1), and shows the ratio of the average speed of vehicles to the speed limit in point $\sigma$ at time $\tau$. Although it is extremely rare that the average speed exceeds speed limit, in cases which exceed, $c_{\tau}^{\sigma}$ is considered as 1.  
It is worth mentioning that Snapshot($\mathcal{S}$,t) is the input unit for predicting the traffic flow of $\mathcal{S}$ at time \textit{t+1}(for example in 20 minutes). In other words, for predicting the traffic flow of $\mathcal{S}$ at time \textit{t+1}, we need recent $\delta$ traffic condition of point $\mathcal{S}$, as well as recent $\delta$ traffic condition of In($\mathcal{S}$), and recent $\delta$ traffic condition of Out($\mathcal{S}$).

\textbf{Definition 5 (Network snapshot).} Assume $\mathcal{N}$ = $\left\{ \mathcal{S}_1, \mathcal{S}_2, \mathcal{S}_3, ..., \mathcal{S}_N\right\}$. Snapshot of $\mathcal{N}$ at time \textit{t}, denoted by SNAPSHOT($\mathcal{N}$, \textit{t}), and defined as follows:

\begin{center}
SNAPSHOT($\mathcal{N}$, \textit{t}) = $\bigcup\limits_{i=1}^{N}$ Snapshot($\mathcal{S}_i,t)$
\end{center}

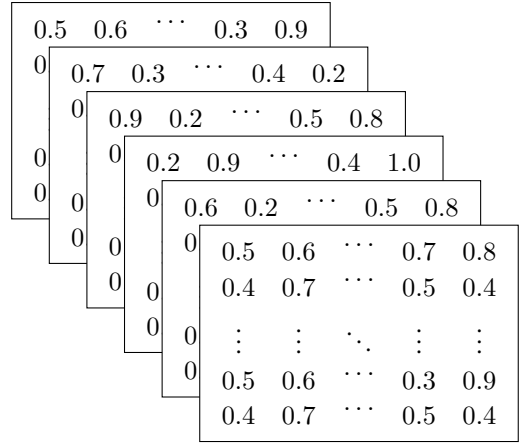
\begin{figure}
\centering
\begin{tikzpicture}[every node/.style={anchor=north west,fill=white,minimum width=0.8cm,minimum height=1.4mm}]
\matrix (mA) [draw,matrix of math nodes]
{
0.5&0.6&\ldots&0.3&0.9\\
0.4&0.7&\ldots&0.5&0.4\\
\vdots &\vdots& \ddots & \vdots & \vdots\\
0.5&0.6&\ldots&0.3&0.9\\
0.4&0.7&\ldots&0.5&0.4\\
};
\matrix (mB) [draw,matrix of math nodes] at ($(mA.south west)+(0.5,2.3)$)
{
0.7&0.3&\ldots&0.4&0.2\\
0.4&0.7&\ldots&0.5&0.4\\
\vdots &\vdots& \ddots & \vdots & \vdots\\
0.5&0.6&\ldots&0.3&0.9\\
0.4&0.7&\ldots&0.5&0.4\\
};

\matrix (mC) [draw,matrix of math nodes] at ($(mB.south west)+(0.5,2.3)$)
{
0.9&0.2&\ldots&0.5&0.8\\
0.4&0.7&\ldots&0.5&0.4\\
\vdots &\vdots& \ddots & \vdots & \vdots\\
0.5&0.6&\ldots&0.3&0.9\\
0.4&0.7&\ldots&0.5&0.4\\
};

\matrix (mD) [draw,matrix of math nodes] at ($(mC.south west)+(0.5,2.3)$)
{
0.2&0.9&\ldots&0.4&1.0\\
0.4&0.7&\ldots&0.5&0.4\\
\vdots &\vdots& \ddots & \vdots & \vdots\\
0.5&0.6&\ldots&0.3&0.9\\
0.4&0.7&\ldots&0.5&0.4\\
};

\matrix (mE) [draw,matrix of math nodes] at ($(mD.south west)+(0.5,2.3)$)
{
0.6&0.2&\ldots&0.5&0.8\\
0.4&0.7&\ldots&0.5&0.4\\
\vdots &\vdots& \ddots & \vdots & \vdots\\
0.5&0.6&\ldots&0.3&0.9\\
0.4&0.7&\ldots&0.5&0.4\\
};

\matrix (mF) [draw,matrix of math nodes] at ($(mE.south west)+(0.5,2.3)$)
{
0.5&0.6&\ldots&0.7&0.8\\
0.4&0.7&\ldots&0.5&0.4\\
\vdots &\vdots& \ddots & \vdots & \vdots\\
0.5&0.6&\ldots&0.3&0.9\\
0.4&0.7&\ldots&0.5&0.4\\
};
%\draw[dashed](mA.north east)--(mC.north east);
%\draw[dashed](mA.north west)--(mC.north west);
%\draw[dashed](mA.south east)--(mC.south east);
\end{tikzpicture}
\caption{Schematic view of SNAPSHOT($\mathcal{N}$, \textit{t})}
\label{fig:snapshotOfN}
\end{figure}

Where Snapshot($\mathcal{S}_i,t)$ is the Point snapshot of $\mathcal{S}_i$ at time \textit{t}, and $\bigcup$ is the union of snapshots of the Network points. Therefore, Snapshot($\mathcal{S}_i,t)$ is a set of all point snapshots of Network points. Fig.\ref{fig:snapshotOfN} schematically present SNAPSHOT($\mathcal{N}$, \textit{t}).

In Fig.\ref{fig:snapshotOfN}, each box consists of the snapshot of one Network point at time \textit{t}, thus all boxes indicate snapshot of the whole traffic network. Assume, Fig. \ref{fig:snapshotOfN} is the current snapshot of the traffic network (SNAPSHOT($\mathcal{N}$, \textit{now})). And, It may be the input for prediction of the next time point (for example, in 20 minutes).
\subsection{Problem Definition}\label{problem}
Assume we have the historical Network snapshots of a region (e.g. North California), gained from traffic detectors located on Traffic points (Definition 1). Our system can learn traffic patterns from this historical data. Suppose, we have the current and the recent Network snapshots of another region (e.g. Southern California), because recently the latter region was equipped with traffic sensors. In this work, we try to predict the traffic flows of the latter region based on the traffic patterns learned from the former traffic network.  

\textbf{Problem.} Given the historical traffic observations of Network $\mathcal{N}_1$ and Snapshot($\mathcal{S}$, \textit{now}), $\mathcal{S} \not\in \mathcal{N}_1$ , predict traffic condition of $\mathcal{S}$ at (\textit{now +} 1), where (\textit{now +} 1) is the next time point (e.g. traffic condition in 20 minutes from now). 

\section{Deep traffic flow network}\label{method} 
In this section, we try to use two deep learning model to solve the problem defined in subsection \ref{problem}. In order to do this, we utilize two prominent algorithms, namely, \textit{Convolutional Neural Network} \cite{krizhevsky2012imagenet} in subsection \ref{conv} and Long Short-Term Memory \cite{hochreiter1997long} in \ref{lstm}.  
\subsection{Deep Traffic Flow convolutional Network}\label{conv}
\begin{figure*}[t!]
\centering
  \includegraphics[width=16cm]{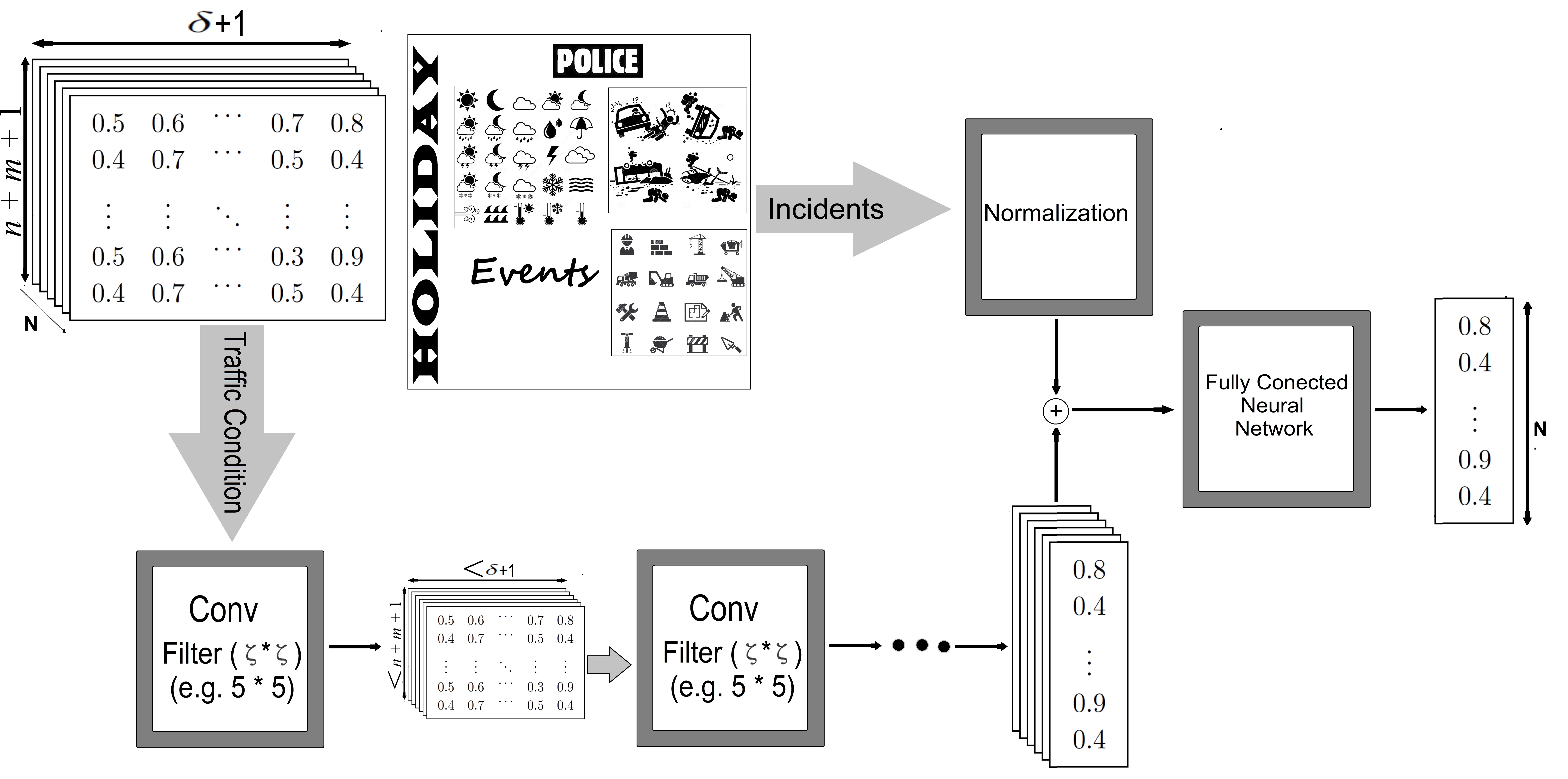}
  \caption{Deep Traffic Flow Convolutional Network}
  \label{Convolutional}
\end{figure*}
Now that we defined all the main concepts about Traffic Flow Prediction, we need to introduce our Deep Traffic Flow convolutional Network. Fig.\ref{Convolutional} illustrates the model while training. As seen, the inputs to this model are categorized in two broad groups, namely, \textit{Traffic Condition} and \textit{Incidents}. \textit{Traffic Conditions} are the set of Network snapshots (See Definition 5). In other words, \textit{Traffic Conditions} is as follows:
\begin{center}
\textit{Traffic Conditions} = $\bigcup\limits_{i=1}^{Z}$ SNAPSHOT($\mathcal{N},t_i)$
\end{center}
Where \textit{Z} is the size of the dataset, chosen for training the model, and SNAPSHOT($\mathcal{N},t_i)$ is the Network snapshot in each training item. In Fig.\ref{Convolutional}, we have \textit{N} points in our Network $\mathcal{N}$. For each Network point at a particular time point, we need to show the snapshot (a 2D array with size (($\delta+1) \times (\textit{n + m +} 1)$). 

On the other hand, \textit{Incidents} are Weather inputs or information about Car accidents, Holiday dates, Road construction, and Events, such as soccer match or concerts. Also, sometimes Police may change the traffic flow. Intuitively, we know these incidents may have huge influence on traffic flow.
In the first step, past \textit{Traffic Conditions} are given to the first layer of Convolutional Neural Network as the training set. Then, the result of first layer is given to the second Convolutional layer. This trend continues until we have a set of one dimensional arrays.  As we know, in each layer of Convolutional, the size of input will decrease (because of the filter ($\zeta * \zeta$) applied on the input). 

When the output of a Convolutional layer is a set of 1D values, it is time to add the incident information to the output, and set the input for Fully-connected Network. The outcome Fully-connected Network outcome is the predicting values of Traffic Flow at \textit{t} + 1.
While training, this output should be compared to the actual traffic flows, and \textit{loss} value should be calculated. By getting help from \textit{loss} value, the \textit{weights} and \textit{bias} values may be updated.
\begin{figure}[t!]
\centering
\includegraphics[width=3.5in]{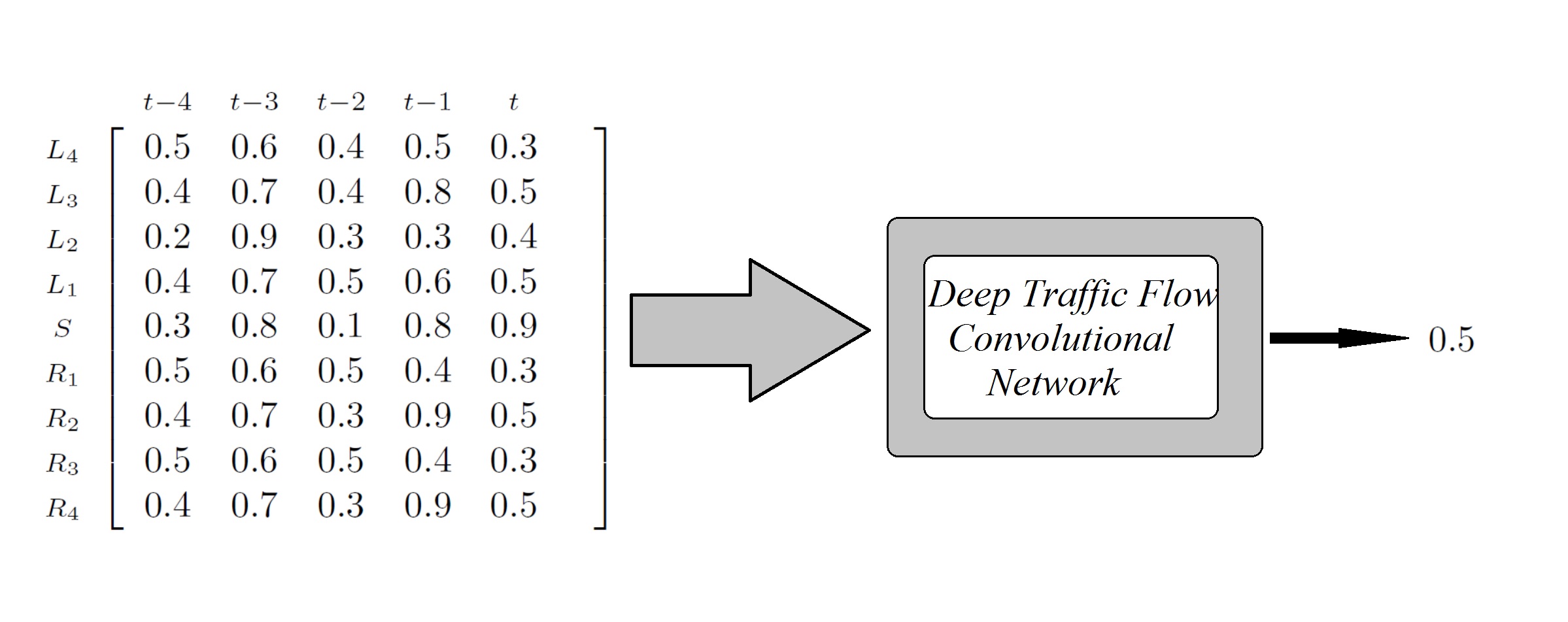}
\caption{Test stage of Deep Traffic Flow Convolutional Network}
\label{fig:conv-test}
\end{figure}
In this work, we try to solve the problem introduced in subsection \ref{problem}. In order to do this, we used the trained aforementioned model. Fig. \ref{fig:conv-test} shows the test model of Fig. \ref{Convolutional}. Having the the traffic condition of  point $\mathcal{S}$ (the point which we try to  predict), and traffic condition of In($\mathcal{S}$) and Out($\mathcal{S}$) in time interval [$now - \delta, now$], the output is the traffic condition of $\mathcal{S}$ at \textit{now} + 1.   
\subsection{Long short-term memory Traffic Flow}\label{lstm}
\begin{figure}[t!]
\centering
\includegraphics[width=3in]{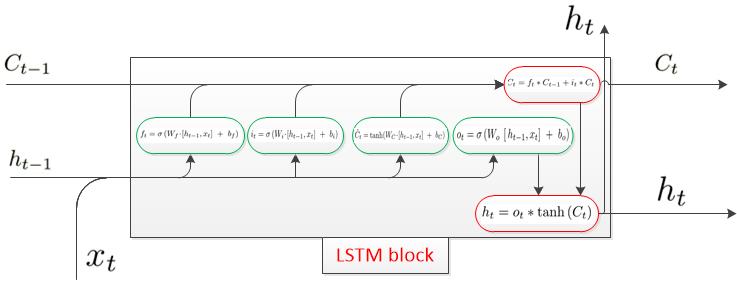}
\caption{An instance of LSTM module}
\label{fig:LSTM}
\end{figure}

In subsection \ref{conv}, we introduce our Deep Traffic convolutional Network for solve the problem defined in section \ref{problem}. In this subsection, we bring up another method, called \textit{Long short-term memory Traffic Flow}. In this method, we use the Long short-term memory (LSTM) network \cite{hochreiter1997long}. LSTM is a Recurrent Neural Network \cite{goller1996learning} with more complex structure in the repeating modules. In other words, each repeating modules contains 4 interacting layers, thus LSTM avoids the long-term dependency problem. Fig. \ref{fig:LSTM} shows an instance of an LSTM module.

\begin{figure}[t!]
\centering
\includegraphics[width=3in]{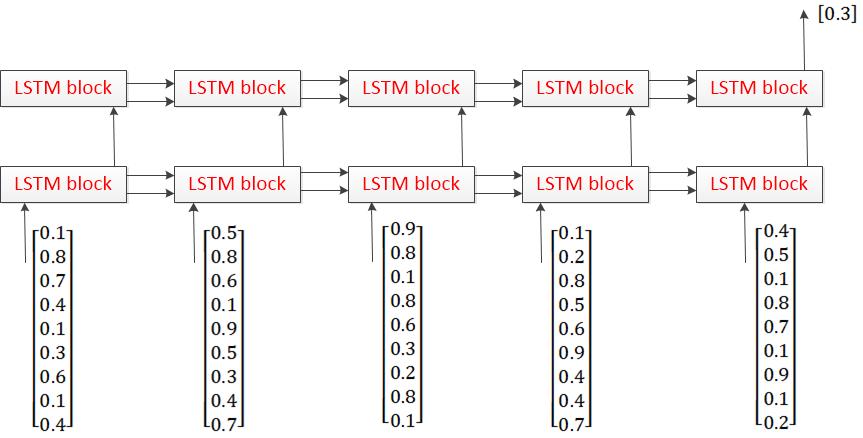}
\caption{Long short-term memory Traffic Flow}
\label{fig:StakcedLSTM}
\end{figure}

In this work, as we mentioned before, we try to predict the short-term future traffic condition for a Network point, where we do not have the historical Traffic condition about it, based on traffic patterns we learned from another road network. Considering the structure of LSTM, our whole network is as in Fig. \ref{fig:StakcedLSTM}. In Fig. \ref{fig:StakcedLSTM}, the traffic condition of  point $\mathcal{S}$ (the point which we try to  predict), and traffic condition of In($\mathcal{S}$) and Out($\mathcal{S}$) at time $t - \delta$, are given to the first LSTM module. Then, in the next step the traffic condition of  point $\mathcal{S}$, In($\mathcal{S}$), and Out($\mathcal{S}$) at time $t - (\delta-1)$, are given to the second LSTM module. This trend continues until the last module. In the last module the output is the traffic condition of $\mathcal{S}$ at \textit{now} + 1.
\section{Experimental Evaluation}\label{exprimental}
In order to examine the performance of our model, we explain our models in detail followed by introducing our dataset. In subsection \ref{Data_set}, we describe our dataset, and in subsection \ref{Deep learning-based}, we bring up our method more in detail.
\subsection{Data set}\label{Data_set}
In all experiments of this evaluation, we used real traffic data of California State. In order to do this, we used the data presented by Caltrans Performance Measurement System (PeMS)\cite{PeMS}. PeMS utilized over 39,000 detectors to collect real-time traffic data. These sensors cover the freeway system across all major metropolitan areas of the State of California. Therefore, PeMS prepared over 10 years of (historical) traffic flow data. The available data in PeMS are not limited to traffic flow data, but it also archived incidents data, such as, Car accidents, Weather information, Lane closures, etc. In this paper, we trained our model for 51 locations on duration of 48 days. Then, we test the model for the same location for the next 12 days. The raw dataset and cleaned dataset are available in \cite{dataset}. These datasets can be used as the benchmark for future Traffic Flow Prediction models.

The data in PeMS are raw data. In order to use them in our model, we did the following pre-processing steps: 
\subsubsection{Getting the data}
The first step is getting the data from \cite{PeMS}. Thus, we download traffic flows (average speeds) of 58 consecutive locations of \textit{US 101} highway for 60 days (each detector on the freeway indicates one location). The aforementioned 60-days is from ``August 15, 2016" to ``October 14, 2016". The granularity for the data is 5 minutes. By way of explanation, we have every-five-minute traffic flow of 58 Network points for 60 days (from ``August 15, 2016" to ``October 14, 2016").
\subsubsection{Cleaning the data}
The next step is the data cleaning. In order to do this, we checked the data and we noticed for a few temporal points, there is not any traffic flow information. For example, we have the traffic flow for $Day_n$ at 13:45 and 13:55, but no data for 13:50. In this case, we calculate the mean values of data at 13:45 and 13:55, and consider them for 13:50.
\subsubsection{Normalizing the data}
 The main part of our data gathering is the \textit{normalizing} of the data. In our experiment, we consider the number of In-flow and Out-flow for each network point are 4 (see Definition 2 and Definition 3). The In-flow and Out-flow values can be defined based on condition of case study, such as, the distance between two consecutive Network Points, average speed, etc. Defining the size of In-flow and Out-flow can be consider as the future work. Also, we assume that in each point snapshot we have traffic conditions of time \textit{t}, and 4 time steps before \textit{t} ($\delta$ = 4, see Definition 4). Therefore, point snapshots are 9 $\times$ 5 (( \textit{n + m +} 1) $\times$ ($\delta$ + 1)). Although we 58 network position, we will not be able to make point snapshots for more than 50 of them, because for the first 4 points we do not have In-flow, and the last 4 of them, we do not have Out-flow.
 
Besides, we batch the data in buckets of 30-minutes based on the day of the week and time. Put differently, all the   traffic flows of Mondays at 8:00 am are considered in the same buckets. Thus, we assign values of ``0 to 6" to the days of the week (i.e. 0 is assigned to Sunday, 1 is assigned to Monday, and so on). At same time we assign values ``0 to 47" to the time of the day (i.e. 0 is assigned to [00:00, 00:30), 1 is assigned [00:30, 01:00), and so on).

Now that we convert time to two values, it is the time to normalize and prepare our data for \textit{training} and the \textit{test}. For normalizing the data, we need to convert all the values (traffic flow information and time) to values in the range [0, 1]. In order to do this, we divide all values by their possible maximum values. For this reason, we divide days and time points by 6 and 47 respectively. Also, we find the ratio of traffic flow of each point to the possible max speed (Speed limit). Hence, a sample of point snapshot is as follows: 

\begin{center}
0.5, 0.6
\[
\begin{array}{lc}
 & \kbordermatrix{\text{}&t-4&t-3&t-2&t-1&t\cr
L_4&0.5&0.6&0.4&0.5&0.3\cr
L_3&0.4&0.7&0.4&0.8&0.5\cr
L_2&0.2&0.9&0.3&0.3&0.4\cr
L_1&0.4&0.7&0.5&0.6&0.5\cr
S&0.3&0.8&0.1&0.8&0.9&\cr
R_1&0.5&0.6&0.5&0.4&0.3\cr
R_2&0.4&0.7&0.3&0.9&0.5\cr
R_3&0.5&0.6&0.5&0.4&0.3\cr
R_4&0.4&0.7&0.3&0.9&0.5} \\[25pt]
\end{array}
\]
\end{center}

In this instance, Day value and Time value are equal to 0.5 and 0.6, respectively (i.e. It is Wednesday at 14:00 to 14:30). In the matrix of traffic conditions, we have 9 rows, indicating the 4 In-flows, the 4 Out-flow, and the Source which is supposed to be predicting. Also, there are 5 columns, 1 column for time \textit{t}, and 4 columns for the last 4 time points before \textit{t}. The time interval between two consecutive time values is 5 minutes (the granularity we chose while getting the data).

\begin{figure*}[htp!]
\centering
\includegraphics[width=16cm]{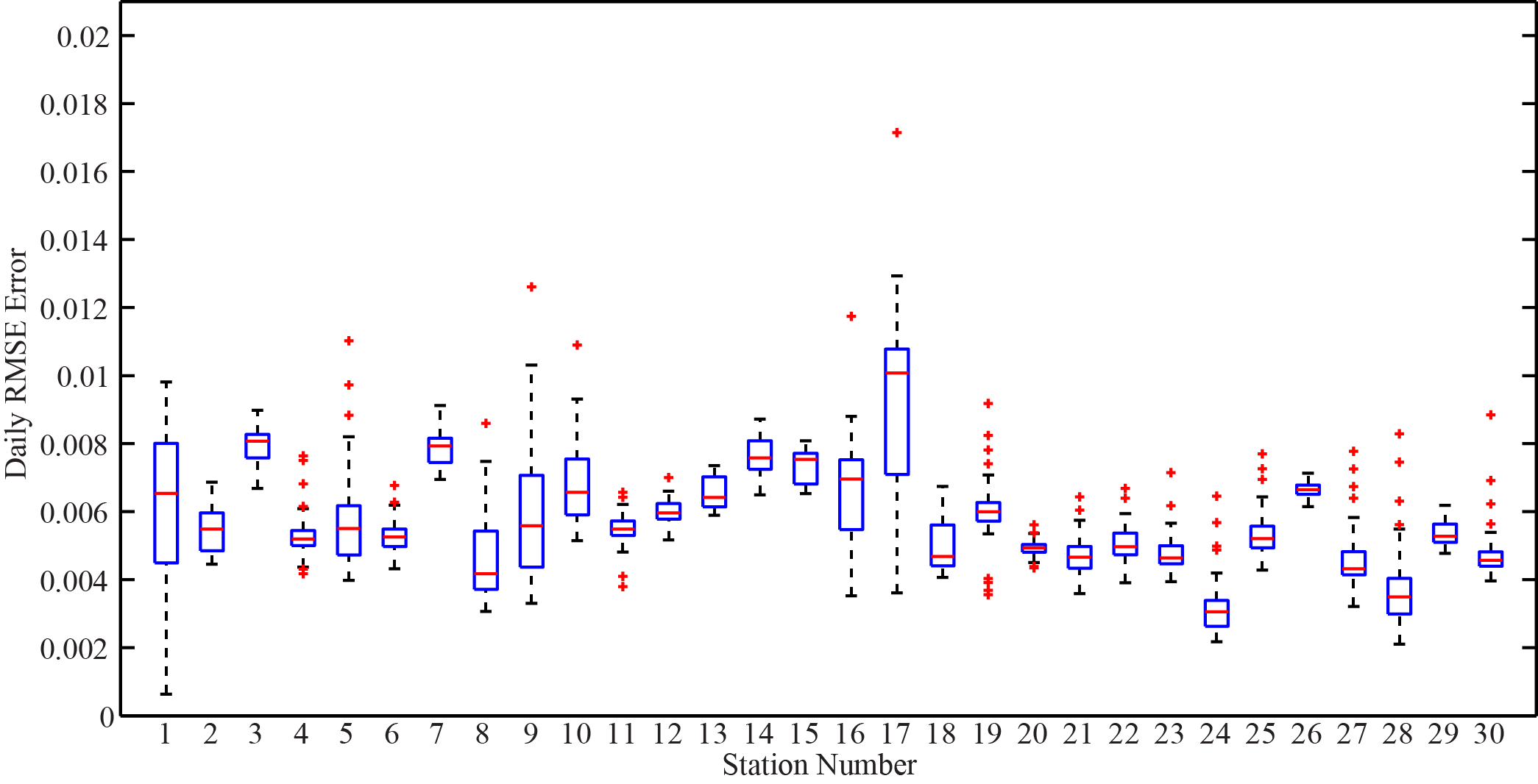}
\caption{Daily RMSE Error  for CNN}
\label{fig:CNN30}
\end{figure*}

\subsection{Deep learning-based method} \label{Deep learning-based}
Now that we have the normalized cleaned data from the subsection \ref{Data_set}, we need to implement two aforementioned methods in section \ref{method}, namely, \textit{Deep Traffic Flow convolutional Network} and \textit{Long short-term memory Traffic Flow} to learn the traffic flow (traffic pattern) of one specific Road Network, $\mathcal{N}_1$). Then, the learned models should be used to predict another road Network, $\mathcal{N}_2$ (see subsection \ref{problem}). In this experiment, we use the first 20 Network points of normalized cleaned data as the training set and the rest 30 Network points as the test set. As we know, former Network points are southern part of dataset, and the latter ones belong to the northern section. Therefore, the two sets are disjoint in order to evaluate the generalization ability of the proposed methods (see subsection \ref{problem}). 

In order to evaluate the proposed methods, we employed a system as follows:

\begin{itemize}
\item Intel(R) Core(TM) i7 CPU 960 @ 3.20GHZ
\item 8GB Ram
\item Quadro 600 NVDIA GPU 1GB
\end{itemize}  

We designed our Deep learning architectures using Lasagne Deep learning framework \cite{sander_dieleman_2015_27878} installed on Theano \cite{2016arXiv160502688short}.

\begin{table}[]
\centering
\caption{Specification of the proposed Convolutional Neural Network}
\label{Conv_table}
\begin{tabular}{|l|l|l|}
\hline
\textbf{Layer Type} & \textbf{Input}           & \textbf{Output}          \\ \hline
Conv1               & 9 $\times$  5            & 7 $\times$ 3 $\times$ 64 \\ \hline
Conv2               & 7 $\times$ 3 $\times$ 64 & 5 $\times$ 1 $\times$ 64 \\ \hline
Fully-Connected     & 320                      & 32                       \\ \hline
Fully-Connected     & 32                       & 1                        \\ \hline
\end{tabular}
\end{table}

The specifications of the proposed Networks are presented in Tables \ref{Conv_table} and \ref{LSTM_table}, where inputs and outputs size of each layer, as well as their filters size are tabulated.  
  
\begin{table}[]
\centering
\caption{Specification of the proposed LSTM network}
\label{LSTM_table}
\begin{tabular}{|l|l|l|l|}
\hline
Layer Type & Input        & Output       & \#Hidden Nodes \\ \hline
LSTM1      & 9 $\times$ 5 & 9 $\times$ 5 & 20             \\ \hline
LSTM2      & 9 $\times$ 5 & 1 & 20             \\ \hline
\end{tabular}
\end{table}

It is worth mentioning that, in this work, we used the Euclidean loss function to train the networks since our proposed problem is intrinsically categorized as Regression. Besides, intuitively, we know that traffic flow data are unbalanced. In other words, traffic flows are typically heavy in few hours of a day. Therefore, our data while traffic was heavy are significantly less than the light traffic data. In order to avoid biased training, we used a regularization equation for our loss as follows:
\begin{equation}
Loss = \frac{\sqrt{\sum_{i=1}^{N} [(X_{i}-Y_{i})^{2} + \omega_{i}\alpha]}}{N}
\end{equation}

where $\alpha$ is the absolute value of $X_i-Y_i$, and $\omega_i$ is as follows:

\begin{center}
\[
    \omega_{i} =  
\begin{cases}
    0,&  if  \hspace{1em}  Y_{i} > 0.5\\
    1,              & \hspace{1em}  otherwise
\end{cases}
\]
\end{center}  

\begin{figure*}[t!]
\centering
\includegraphics[width=16cm]{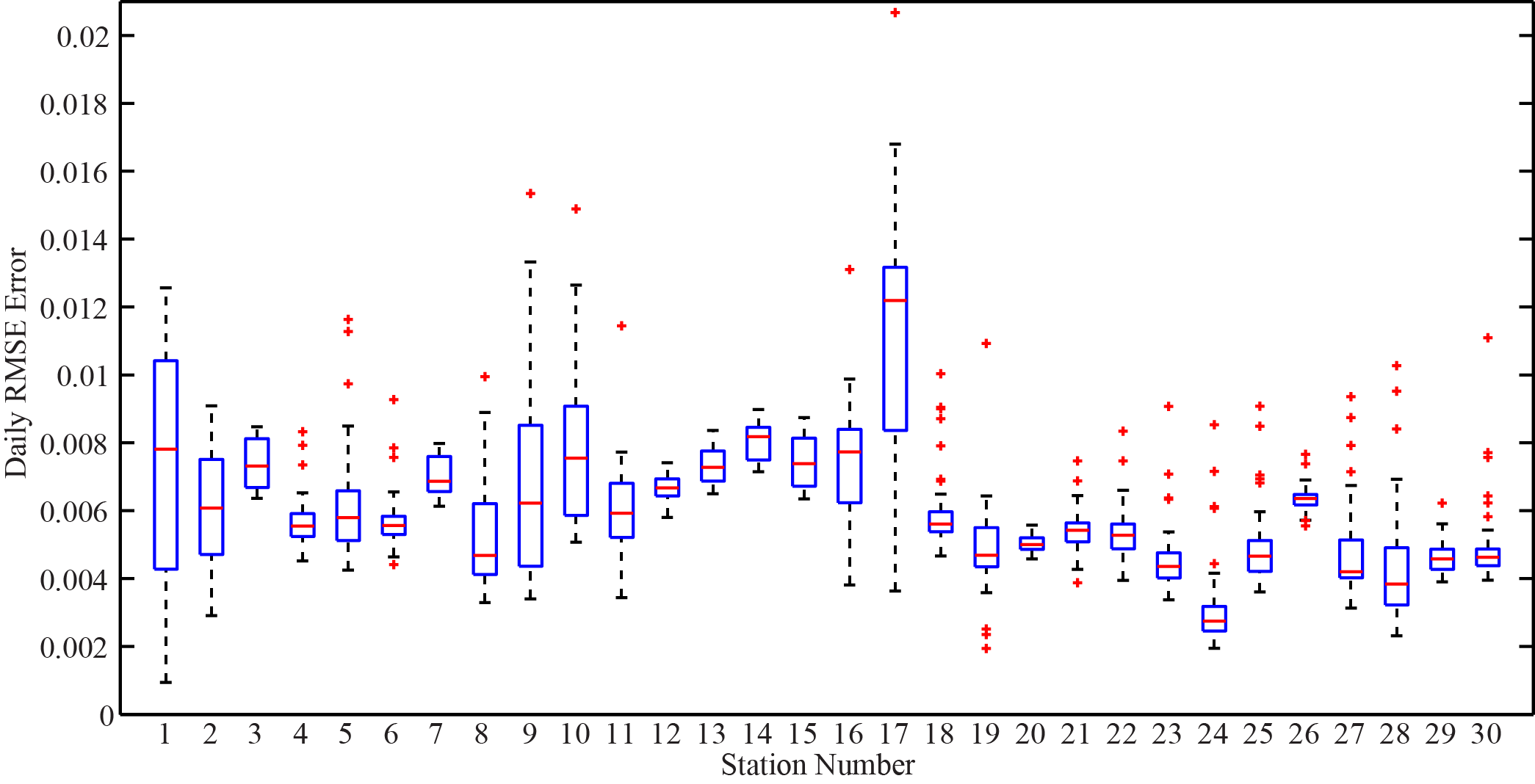}
\caption{Daily RMSE Error for LSTM}
\label{fig:LSTM30}
\end{figure*}

Each of the proposed networks are trained for 30 epochs over the trained dataset containing the flow information for 20 Network points over of the course of 2 months. The trained networks are then used to predict congestion conditions of the road network.

Fig. \ref{fig:CNN30} illustrates the Root Mean Square error (RMSE) error for 30 Network points. The RMSE is computed for each day at a specific Network point and the distribution of RMSE error for each Network point is plotted as a box plot in Fig. \ref{fig:CNN30} for the CNN. On the other hand, Fig. \ref{fig:LSTM30} presents the same RMSE plot for LSTM network. As these plots illustrate, the RMSE for CNN is lower than LSTM and thus results in lower standard deviation. 

\begin{figure}[t!]
\centering
\includegraphics[width=3.in]{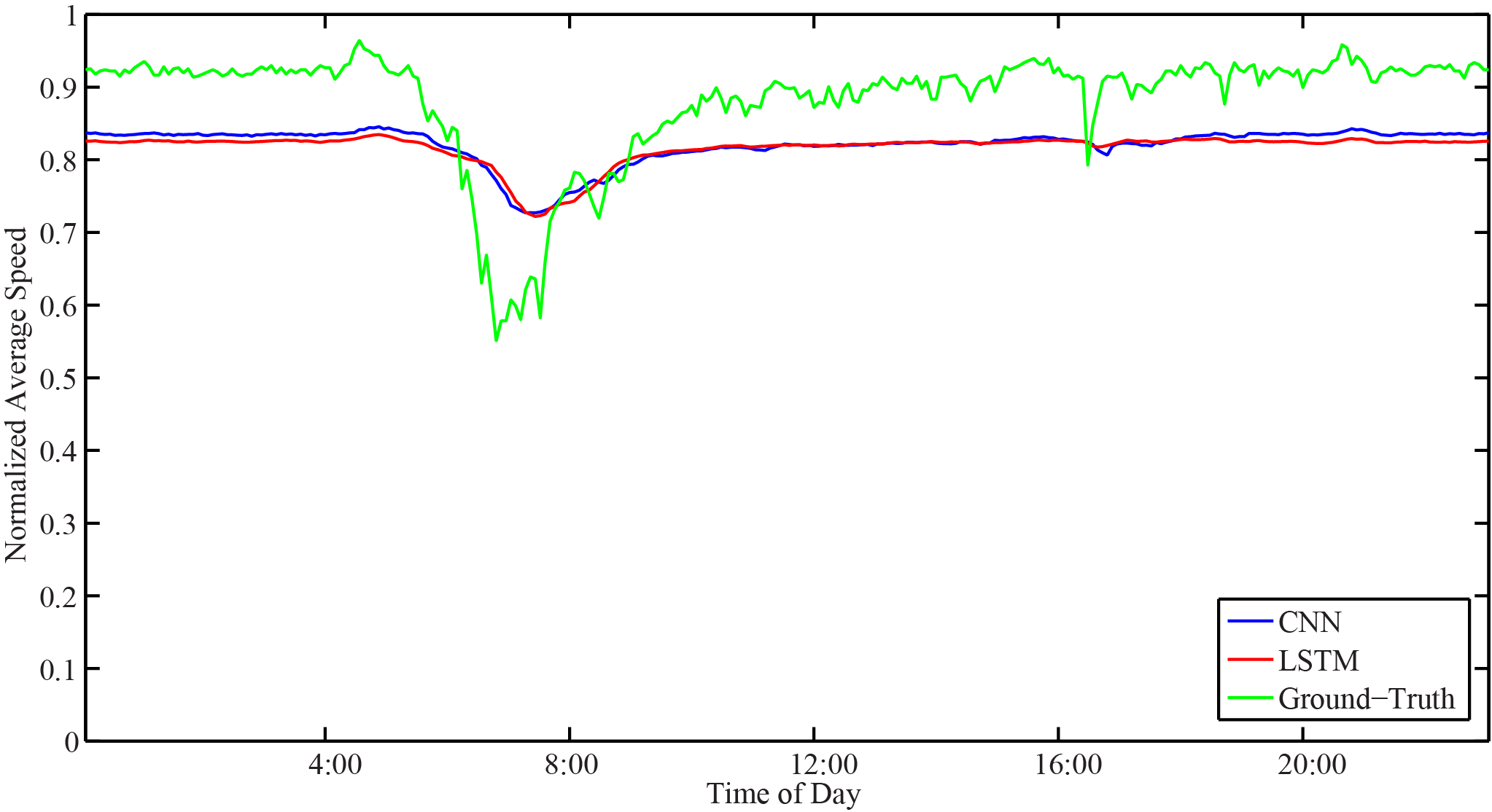}
\caption{Traffic Prediction over the course of a day for a single Network point}
\label{fig:monday30}
\end{figure}

In order to further evaluate the performance of the proposed methods on the benchmark, we examine the prediction of each network over the course of a day. Fig. \ref{fig:monday30} presents the predicted normalized average speeds for CNN and LSTM networks as well as the ground-truth data. The proposed methods successfully predict the congestion condition with high accuracy in compared with the ground-truth data. However, the error increases slightly during the rush hours and that is due to the unbalanced nature of the dataset. However, The proposed regularization term effectively decreased the gap during the rush hour.

\begin{figure}[t!]
\centering
\includegraphics[width=3.5in]{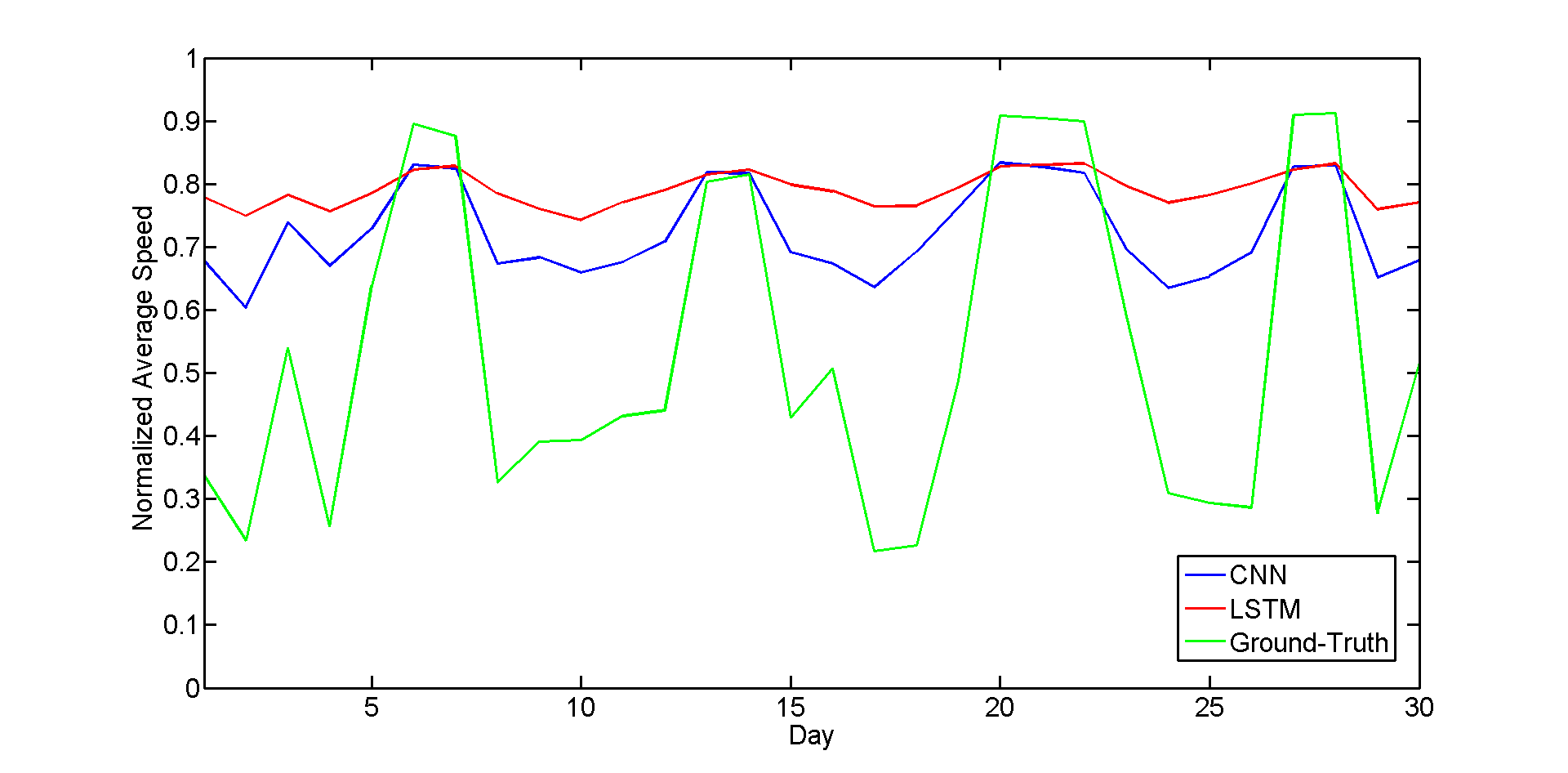}
\caption{Predicted traffic flow during rush hours in 30 consecutive days for a single Network point}
\label{fig:RushHour30Days}
\end{figure}

Hereby, we take a deeper look into the prediction during the rush hours. Fig. \ref{fig:RushHour30Days} illustrates the congestion prediction for a single Network point over 30 consecutive days. For this experiment we utilized the information for Network point number 21 at 7:30 am. As shown in this figure, the error increases as the congestion increases however, both networks successfully follow the trends.

\begin{figure}[t!]
\centering
\includegraphics[width=3.5in]{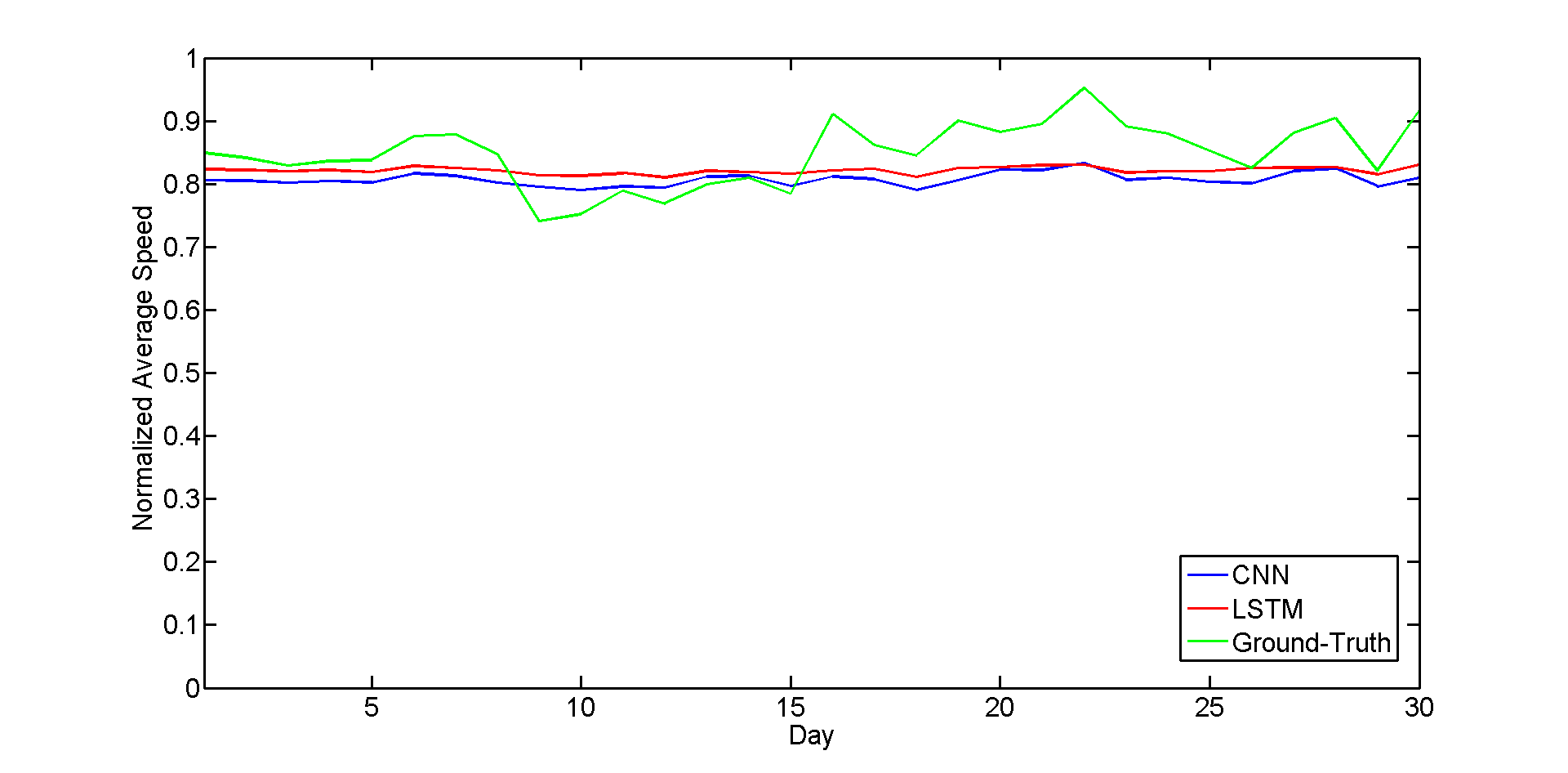}
\caption{Predicted traffic flow during light-traffic hours in 30 consecutive days for a single Network point}
\label{fig:NonRushHour30Days}
\end{figure}

On the contrary, Fig. \ref{fig:NonRushHour30Days} illustrates the predicted network flow using LSTM and CNN for light-traffic conditions where we picked 12:00 pm as an example to plot the predicted average speed for Network point number 21. As shown in this figure, the predictions are more accurate in this case.

\section{Conclusion}
Concepts of traffic bottleneck and congestion propagation are critical components of Intelligent transportation network management systems. There have been lot of effort to understand how the traffic flows and short-term prediction of congestion occurrence because of rush hours or incidents, such as car crashes or Sport events, can be beneficial to such systems to effectively manage and direct the traffic to the most appropriate detours. Most of traffic flow prediction systems rely on utilizing a central processing component where the prediction is carried out through aggregation of the information gathered from all measuring stations. Nevertheless, such system are typically scalable and unable to provide real-time feedback to the system whereas in a decentralized scheme, each node is responsible to predict its own short-term congestion based on the local current measurements in neighboring nodes.

In this work, we introduced a scalable decentralized traffic flow prediction by utilizing deep learning-based method. Therefore each node accurately predicts its own congestion state in real-time based on the congestion state of the neighboring Network point. Besides, proposed method is significantly suitable in the cases, where we need to predict the traffic flow of newly installed stations, using Deep Network trained by historical data from another traffic network. we introduced a regularized euclidean loss function that favors high congestion samples over low congestion samples to avoid the impact of the unbalanced training dataset. A novel dataset for this purpose was designed based on the traffic data obtained from traffic control stations in northern California. Extensive experiments conducted on the designed benchmark reflected a successful congestion prediction.

\bibliographystyle{abbrv}
\bibliography{bare_conf}

% that's all folks
\end{document}